\documentclass{egpubl}
\usepackage{eurovis2018, amsmath, amssymb}

\DeclareMathOperator*{\argmin}{arg\,min}

\SpecialIssuePaper         
\ifpdf \usepackage[pdftex]{graphicx} \pdfcompresslevel=9
\else \usepackage[dvips]{graphicx} \fi

\PrintedOrElectronic

\usepackage{t1enc,dfadobe}
\usepackage{amsmath,amssymb}
\usepackage{egweblnk}
\usepackage{cite}

\usepackage{algorithmic}
\usepackage[tight,footnotesize]{subfigure}
\usepackage{xargs}                      
\usepackage[pdftex,dvipsnames]{xcolor}  

\usepackage{algorithmic}

\usepackage{array}
\newcolumntype{P}[1]{>{\centering\arraybackslash}p{#1}}
\newcolumntype{M}[1]{>{\centering\arraybackslash}m{#1}}

 \newtheorem{lemma}{Lemma}[section]

 \newtheorem{algorithm}[lemma]{Algorithm}

\graphicspath{{figs/}{pictures/}{images/}{./}} 

\title[Exploring High-Dimensional Structure via Axis-Aligned Decomposition of Linear Projections]
      {Exploring High-Dimensional Structure via Axis-Aligned Decomposition of Linear Projections
      \vspace{-6mm}}

\author [Jayaraman J. Thiagarajan et. al.] {Jayaraman J. Thiagarajan$^{1}$\thanks{This  work  was  performed  under  the  auspices  of  the  U.S.
		Dept. of Energy by Lawrence Livermore National Labora-
		tory under Contract DE-AC52-07NA27344.}, Shusen Liu$^{1}$, Karthikeyan Natesan Ramamurthy$^{2}$, Peer-Timo Bremer$^{1}$ \\
$^{1}$Lawrence Livermore National Laboratory, $^{2}$ IBM T.J. Watson Research Center}

\begin{document}

\teaser{
\vspace{-8mm}
 \includegraphics[width=0.78\linewidth]{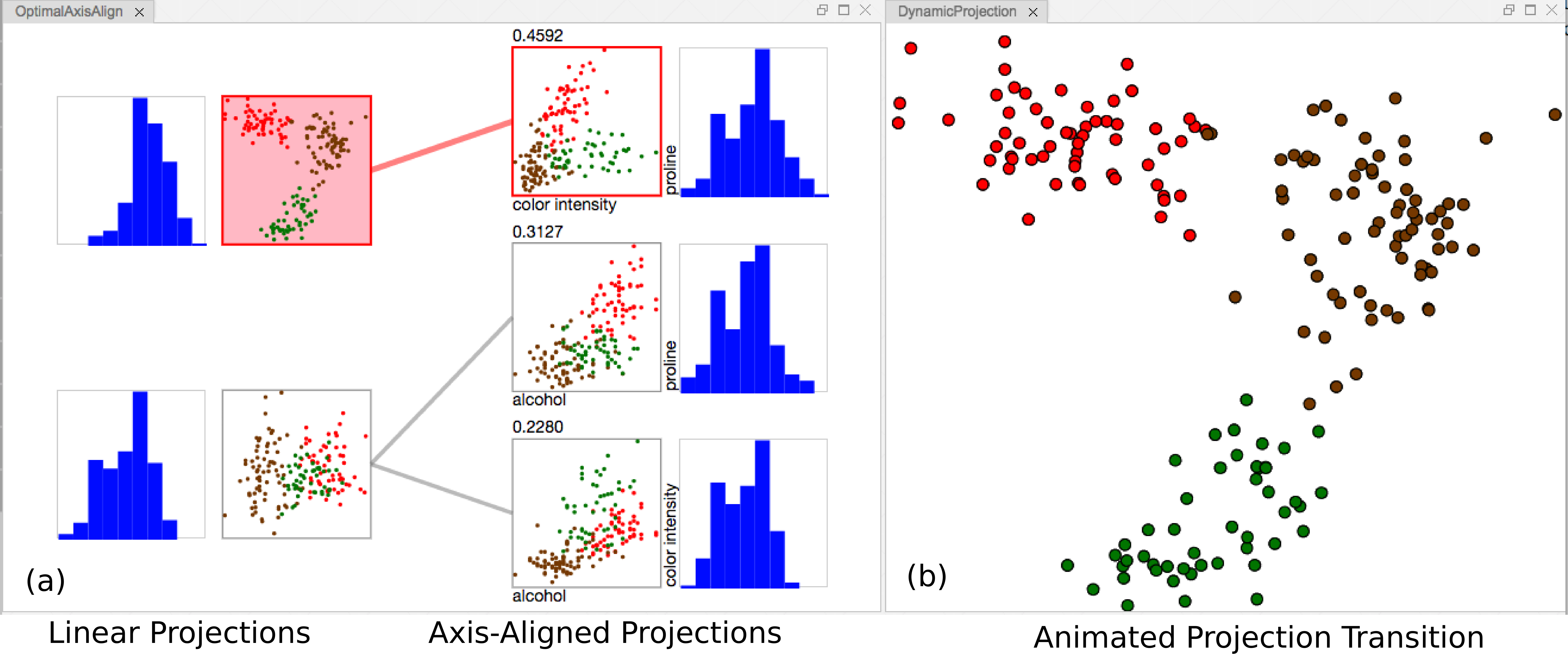}
 \centering
  \caption{Visual interface overview. The proposed method is integrated as part of an existing 
  high-dimensional data visualization system. In the left panel, the high-dimensional data is first
  visualized as a list of representative linear projections and then as axis-aligned projections
  that are a sparse decomposition of the linear ones.
  To illustrate the relationship between the linear and the corresponding
  axis-aligned projections, we show their connections (a) as well as
  animated transitions between projections (b).
  }
\label{fig:teaser}
}

\maketitle

\begin{abstract}
Two-dimensional embeddings remain the dominant approach to visualize
high dimensional data. The choice of embeddings ranges from highly
non-linear ones, which can capture complex relationships but are
difficult to interpret quantitatively, to axis-aligned projections,
which are easy to interpret but are limited to bivariate relationships.
Linear project can be considered as a compromise between complexity and
interpretability, as they allow explicit axes labels, yet provide
significantly more degrees of freedom compared to axis-aligned
projections. Nevertheless, interpreting the axes directions, which are
linear combinations often with many non-trivial components, remains
difficult. To address this problem we introduce a structure aware
decomposition of (multiple) linear projections into sparse sets of
axis aligned projections, which jointly capture all information of
the original linear ones. In particular, we use tools from
Dempster-Shafer theory to formally define how relevant a given axis
aligned project is to explain the neighborhood relations displayed
in some linear projection. Furthermore, we introduce a new approach
to discover a diverse set of high quality linear projections and
show that in practice the information of $k$ linear projections is
often jointly encoded in $\sim k$ axis aligned plots. We have integrated
these ideas into an interactive visualization system that allows
users to jointly browse both linear projections and their axis
aligned representatives. Using a number of case studies we show how
the resulting plots lead to more intuitive visualizations and new
insight.



\end{abstract}


\section{Introduction}
With the ever-increasing emphasis on data-centric analysis, studying
high dimensional data has become an ubiquitous problem in science and
engineering. Traditional confirmatory data analysis~\cite{tukey1980we}
(i.e., confirm/reject an hypothesis) requires users to form
relevant hypotheses before any analysis can be started. However,
deriving an intuitive understanding of the data in order to form such
hypotheses is becoming increasingly difficult. One common approach is
to use (interactive) visual exploration to inspire
new hypotheses and a large body of research has been focused on how to
make this process as intuitive and effective as possible.

While there exist a wide range of different approaches,
two-dimensional embeddings remain the most commonly used technique to
explore high dimensional data. The goal is to allow the user to reason
about high-dimensional relationships and structures, i.e.,
correlations, clusters, etc., in a low dimensional, less abstract
context. To aid the user in obtaining new insights,
these embeddings need to be both accurate, i.e.,
reflect the high-dimensional structures, and
intuitive, i.e., allow for an easy interpretation. Unfortunately, these
two goals often conflict in practice. On one end of the spectrum
non-linear embeddings, e.g., t-distributed stochastic neighborhood embedding (t-SNE)~\cite{MaatenHinton2008} and multi-dimensional scaling (MDS)~\cite{KruskalWish1978}, are
good at capturing complex relationships. However, the resulting
embeddings are difficult to interpret quantitatively as directions in
the plot do not necessarily correspond to any coordinate, and distances
can be severely distorted. Conversely, axis-aligned projections are
straightforward to interpret, yet are very limited in the type of
relationships they can highlight.

Linear projections are often considered a good compromise between both
objectives as they can arbitrarily increase the degrees of freedom to find
high dimensional relationships, while directly relating directions in the plot to the original coordinates. Though the distances in a linear projection are easy to interpret, the axes often correspond to linear combinations of dozens of
dimensions, many of which are expressed equally strong. Consequently, this requires a user to simultaneously reason about many different attributes which is often overwhelming. One common approach aiming to address this
challenge is to enforce sparsity in both axes of a plot by preferring linear
projections with fewer constituent coordinates~\cite{Chipman2005}. However, this reduces the expressive power of the resulting linear projections and unless taken to the extreme often still results in too many relevant coordinate directions to be intuitive.

An attractive alternative is to consider a given linear projection in
light of several similar axis-aligned projections. However, the
straightforward solution of constructing a scatterplot matrix from
all (significant) constituent coordinate directions typically results
in too many combinations to be practical. Furthermore, there is little
control over how accurate the resulting plots represent the true
structure and many projections might be misleading rather than
helpful. Instead, we introduce a new approach to approximate a linear
projection with a small number of axis-aligned projections via
generalization of sparse representations to the Riemannian space of linear projections~\cite{TroppGilbert2007}, coupled with a greedy dimension selection technique.
Furthermore, independent of how well the user can interpret a single linear
projection, as corroborated by several recent related works~\cite{NamMueller2014, LiuWangThiagarajan2015, wang2017subspace, lehmann2016optimal, LiuBremerJayaraman2016}, an effective exploration of most high dimensional data requires a diverse set of views. To this end, we extend the decomposition approach to the case of multiple linear projections
with additional constraint to reduce duplicated axis-aligned presentation. In particular, we define a measure of \emph{evidence} using Dempster-Schafer theory to convey how much of the information from multiple linear projection is explained by a certain axis-aligned one. In addition, we introduce a new optimization approach to construct multiple linear projections that are both accurate and diverse. Unlike existing approaches, we do not focus exclusively on diversity~\cite{lehmann2016optimal}, which may lead to projections with poor embedding quality, nor do we require a dense sampling of all possible projections~\cite{LiuBremerJayaraman2016}, or expensive subspace clustering computation~\cite{NamMueller2014, LiuWangThiagarajan2015, wang2017subspace}. Instead, we introduce an iterative optimization to explore the Grassmannian (i.e., the space of all linear projections) with any convex embedding objective, e.g., PCA (principal component analysis), LPP (locality preserving projection). We achieve the co-optimization of both diversity and quality, while still maintaining the convexity for each iteration to solve the problem efficiently.

Finally, we integrate our approaches into an interactive visual analytics interface that allows users to easily and intuitively interpret high
dimensional data through a multi-faceted lens of linear and axis-aligned
projections. Our contributions in detail are:
\begin{itemize}

\item A mathematical framework for representing a linear subspace as a
  sparse set of axis-aligned subspaces in a structure-aware manner;

\item An optimization algorithm for identifying a diverse set of
  linear projections by simultaneously maximizing the accuracy and
  diversity of the projections;

\item A visualization tool that exploits benefits of both the linear
  and axis-aligned projections by summarizing the relationships
  between the selected diverse set of linear projections and their
  corresponding axis-aligned projections.

\end{itemize}

\section{Related Work}
Generating low-dimensional embeddings of high-dimensional data is an extensively
studied area in many related fields such as data mining, machine learning,
and visualization.
%
%
In the visualization community, instead of focusing solely on the general
objective of dimensionality reduction, considerable efforts have been devoted
to user-driven exploration and interpretation of high-dimensional data.
In this section, we will discuss related approaches that rely on both axis-aligned and linear projections for exploratory analysis.

\subsection{Axis-Aligned Projection}
Scatterplot matrix (SPLOMs) is one of the most popular methods for visualizing
high-dimensional data, where each plot is an axis-align projection.
However, due to the quadratic increase in the number of plots as
dimension grows, several methods (e.g., \emph{Scagnostics}
\cite{WilkinsonAnandGrossman2005} and rank-by-feature
framework~\cite{SeoShneiderman2004}) have been proposed to help identify
"interesting" plots that are worthy of the user's attention.
\emph{Scagnostics} \cite{WilkinsonAnandGrossman2005},
supplies the user with multiple types of measures, e.g., \emph{clumpy},
\emph{skinny}, each capturing a specific pattern in the scatterplot.
Combined, the different measures help user select a diverse set of axis-aligned
projections. For an extensive review of the various quality measures, please
refer to the survey articles~\cite{BertiniTatuKeim2011, LiuMaljovecWang2017}.

Compared to the well-known approaches that rely on ranking quality measures,
the proposed approach is fundamentally different. Instead of filtering directly
on all axis-aligned projections, we search for structures in the space of linear
projections and then use axis-aligned projections to help explain the structure observed in them. In other words, linear projections are used as a guide for discovering diverse views of the data and as a bridge to connect the high-dimensional space with axis-aligned projections. For data with relative high-dimensions, the linear projection stage also helps avoid the quadratic complexity of the scatterplot matrix.

\subsection{Linear Projection}
Many widely adopted dimension reduction methods, such as principal component
analysis~\cite{AbdiWilliams2010} and Fisher's discriminant analysis~\cite{MikaRatschWeston1999},
are linear in nature.
These methods are often easy to compute and capture structures that are of great
interest for analysis (e.g., linear correlation, clusters in subspaces).
However, a single linear projection only provides a limited view
of the data, and hence might not reveal other important structures.
As a result, several visualization approaches have resorted to creating
multiple 2D projections or a series of projections (often referred to as a tour)
to provide a more comprehensive view of the data.
The classical \emph{Grand Tour} method~\cite{asimov1985grand} visualizes data through a series of 2D projections that explore the space of all
linear projections in a space-filling manner. However, due to the size
such a space, a complete tour may not be possible, even in moderate dimensions.

Instead of aiming for a complete tour, an alternative approach is
to devise a measure to identify ``interesting'' projections.
In the \emph{projection pursuit}, we seek for 2D projections by
optimizing a quality measure function~\cite{friedman1974projection}.
One example is the \emph{Holes} measure~\cite{CookBujaCabrera1993}, which
finds projections where there is a gap between two clusters of points.
For a given measure, the global extrema provides the user with only one view of the data. In order to address this limitation, the Grassmannian Atlas framework~\cite{LiuBremerJayaraman2016}, adopts tools from topological data analysis, in lieu of the global optimization, and identifies multiple local extrema of the quality measure, thus identifying a complementary set of linear projections.
Recently, subspace clustering/selection based methods~\cite{LiuWangThiagarajan2015,
Tatu2012, NamMueller2014, Yuan2013, wang2017subspace}, have enabled
structure-driven exploration, particularly while identifying important linear projections. Broadly, these methods decompose the high-dimensional data into lower-dimension subsets and project them separately to focus on local features.
Besides using structure/pattern based quality measures to determine ``interesting'' projections, one can also explicitly consider diversity as an objective in the view optimization process. For example, Lehmann et al.~\cite{lehmann2016optimal} proposed to construct a set of 2D projections that are most dissimilar to each other after accounting for rotation and translation of the views. However, by not considering the quality of the resulting embeddings, this approach can produce views that do not strongly agree with the structure in high dimensions.
Compared to the existing methods, the proposed algorithm
considers both the quality, i.e., how well the current projection preserve
neighborhood structure, and diversity, i.e., maximal separation from the
previously selected ones, of the projections.

Reasoning about the meaning of the axes that are expressed as linear combinations of many different properties can be very challenging.
Existing methods~\cite{Morton1992, Chipman2005} aim to reduce the number
of non-zero components in the linear combination. 
In particular, the interpretable dimensionality reduction technique~\cite{Chipman2005} employs
an explicit sparse constraint to the linear projection bases, which yield simpler coordinates for the analysis.
However, these techniques do not resolve the fundamental burden of interpreting
axes as linear combinations, which can be particularly challenging in high dimensions. In contrast, the proposed method provides a set of axis-aligned projections that are inherently easier to interpret, while being maximally descriptive of the linear projections that guided their selection.

\section{Axis-Aligned Decomposition of a Linear Projection}
\label{sec:decomposition}

\begin{figure}[htbp]
\centering
\vspace{-2mm}
 \includegraphics[width=.8\linewidth]{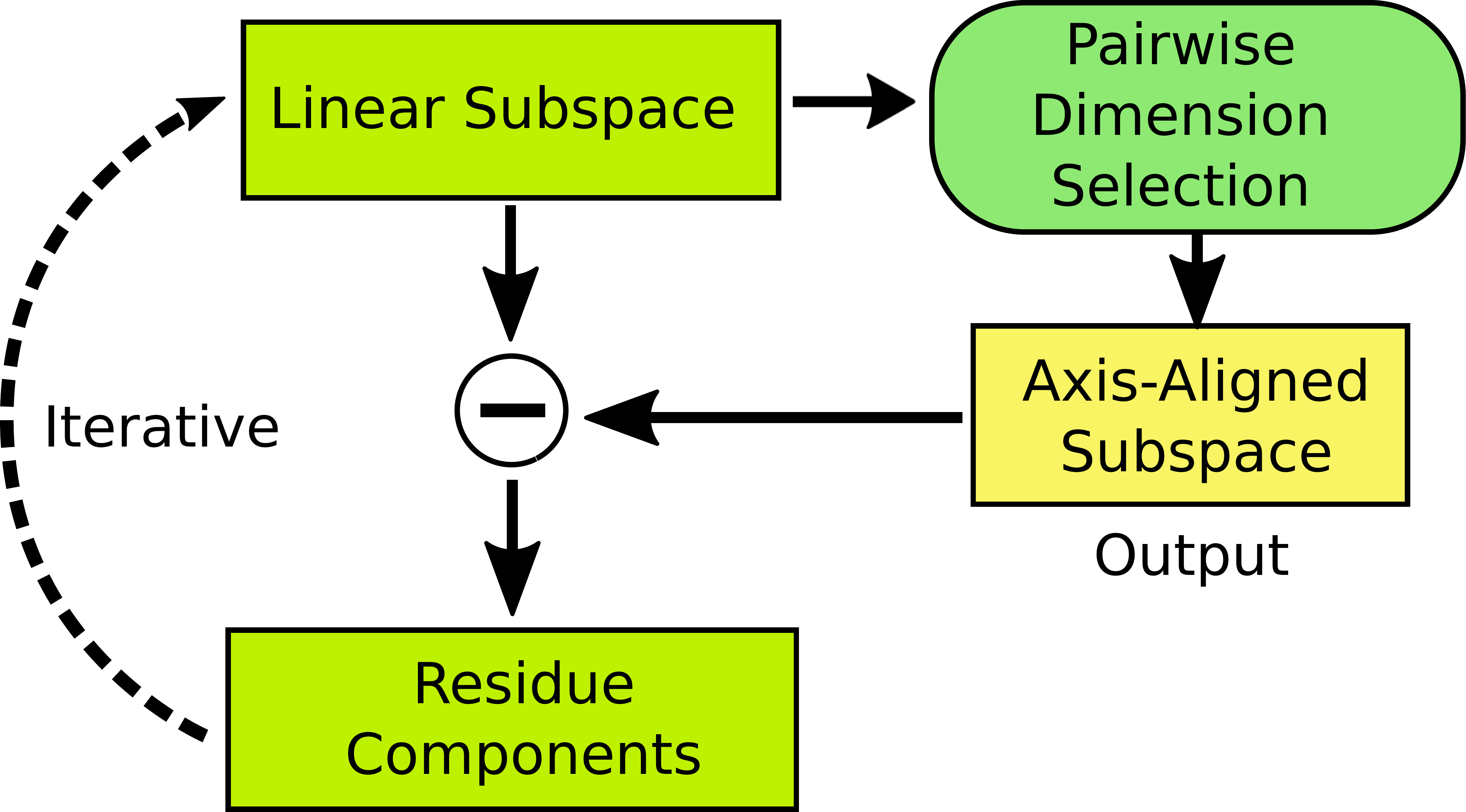}
 \caption{Illustration of the linear projection decomposition algorithm.
 Given a linear subspace, we iteratively find a best matching
axis-aligned one, remove its contribution from the linear projection 
and continue until no additional axis-aligned subspace can provide better structure preservation 
than the ones picked so far.
 }
\label{fig:OMP}
\vspace{-5mm}
\end{figure}

As discussed above, the overarching goal is to find a set of
axis-aligned projections that jointly represent a linear projection
well. A naive approach might use all pair-wise combinations of all
dimensions that are active in a given linear projection. However, in
practice, this will likely result in an overwhelming number of
scatterplots. A more sophisticated variant of this idea is to use
sparse coding on the Grassmannian manifold (the space of all linear
subspaces)~\cite{harandi2013dictionary}, which can be used to find a small set
of axis-aligned projections. More specifically, let $\mathbf{V}$ be a
basis of the linear subspace one is interested in and $\mathcal{Q}$ be
the index set of all pair-wise combinations of dimensions,
$|\mathcal{Q}| = {d \choose 2}$, then to find the $L$ best axis-aligned
subspaces, one can used the formulation of Harandi \textit{et al.}~\cite{harandi2013dictionary}:
\begin{align}
\nonumber
  & \boldsymbol{{\beta}} = \argmin_{\{\beta_i\}} \|\mathbf{V} \mathbf{V}^T - \sum_{i=1}^{|\mathcal{Q}|} \beta_i \mathbf{Z}_i \mathbf{Z}_i^T\|_F^2,\\
  \label{eqn:grassmann_sparse} & \text{subject to }
                                 \|\boldsymbol{\beta}\|_0 \leq L.
\end{align}
\noindent Here $\mathbf{V} \mathbf{V}^T$ is the extrinsic
representation for a point (a linear subspace) $\mathbf{V}$ on the
Grassmannian and $\mathbf{Z}_i$ are axis-aligned subspaces. Note that
this optimization is data independent, i.e., the error considers only
the distance on the Grassmannian. This is not ideal for two reasons:
First, the chosen $\mathbf{Z}_i^{\text{'s}}$ may result in poor projections either because they create significant distortions or their structure is not
relevant to $\mathbf{V}$; and second, some of the $\mathbf{Z}_i^{\text{'s}}$ may
result in very similar and thus redundant projections. The latter is
common in data sets with highly correlated dimensions. Consider the
extreme case of duplicated pairs of dimensions, which result in
identical projections yet are maximally far apart on the Grassmannian.
Instead, we propose to explicitly look for axis-aligned projections
that encode the same structure as the given linear one. More
specifically, we follow the approach shown in Figure~\ref{fig:OMP}:
Given a linear subspace, we iteratively find a best matching
axis-aligned one, remove its contribution from the linear projection on the Grassmannian to estimate the residual subspace and continue until no additional axis-aligned subspace can provide better structure preservation than the ones picked so far.

\subsection{Algorithm}
We are interested in preserving the structure of a linear projection
which we define as the set of all pairwise distances. Given a set of
$n$ points $\mathbf{X}$ in $\mathbb{R}^d$ and the basis of a linear
subspace $\mathbf{V}$, the projected coordinates $\mathbf{Y}$ are given
as $\mathbf{Y} = \mathbf{V}^T \mathbf{X}$. The structure we want to
preserve are pairwise distances $||\mathbf{y}_i - \mathbf{y}_j||^2$
for all $j,i \in {1,...,n}$, using a pair of dimensions from $\mathbf{X}$. In practice, we know that preserving all
distances accurately is infeasible and thus we typically
restrict the distances to a set of nearest neighbors, i.e.,
$j \in\mathcal{N}_k(i)$, where $\mathcal{N}_k(i)$ defines the
$k$-nearest neighbors of $\mathbf{y}_i$ computed using
the subspace defined by $\mathbf{V}$.

In order to find a pair of dimensions that best preserves the neighborhood
distances we modify the image masking technique
of~\cite{dadkhahi2014masking}. Note that, for a uniformly distributed set of random masks of dimension $2$, i.e., can use only two of the $d$ dimensions in $\mathbf{X}$, one can expect a compaction factor of $\sqrt{{2}/{d}}$ in the true distances from $\mathbf{X}$. However, in our case, it is sufficient for the selected dimensions to agree with the structure of $\mathbf{Y}$. Hence, we start by defining the matrix $\mathbf{C}$ by stacking the set of difference vectors squared elementwise, $[(\mathbf{x}_i^1 - \mathbf{x}_j^1)^2,...,(\mathbf{x}_i^d - \mathbf{x}_j^d)^2]$, referred to as unnormalized secants, for all neighboring $i,j$ pairs. Similarly, we construct the unnormalized secants relative to the projection $\mathbf{V}$ and compute vector $\mathbf{b}$ with $\|\mathbf{y}_i - \mathbf{y}_j\|^2$. The optimal axis-aligned projection can then be found by optimizing:
\begin{align}
\nonumber
& \boldsymbol{{\alpha}} = \argmin_{\boldsymbol{\alpha}} \| \mathbf{C} \boldsymbol{\alpha} - \mathbf{b} \|_2^2 \\
\label{eqn:iden_ind}
& \text{subject to} \qquad  \mathbf{1}^T \boldsymbol{\alpha} = 2, \boldsymbol{\alpha} \in \{0,1\}^{d}.
\end{align}

The non-zero entries in $\boldsymbol{{\alpha}}$ then indicate the two
dimensions of the axis-aligned projection that best preserves the
nearest-neighbor distances. Rather than solving this optimization
directly as a binary integer program, we use a greedy procedure that selects one
dimension of $\mathbf{X}$ at a time. The mask vector
$\boldsymbol{{\alpha}}$ has a one-to-one correspondence with the index
set $\mathcal{Q}$, and let us denote the corresponding index by
$i_{{\alpha}}$. The axis-aligned subspace obtained is denoted by
$\mathbf{Z}_{i_\alpha} \in \mathbb{R}^{d \times 2}$.  We can now
describe the full algorithm as shown below:

\begin{algorithm}
  \label{alg:sparse_grass}
  Optimize for $\Omega$ given $\mathbf{V}$ and $L$ (maximum number of axis-aligned subspaces)
  \begin{enumerate}

  \item Initialize axis -aligned projections $\Omega = \varnothing$,
   subspace $\mathbf{U} = \mathbf{V}$

  \item While $|\Omega| \leq L$:

    \begin{enumerate}
    \item Compute optimal $\mathbf{Z}_{i_{{\alpha}}}$ to approximate $\mathbf{U}$ using (\ref{eqn:iden_ind})
    \item Measure structural distortion $\|\mathbf{e}^i = \mathbf{C}\boldsymbol{\alpha} - \mathbf{b}^i\|_2$ for $\mathbf{U}$ and $\mathbf{Z}_{i_{{\alpha}}}$
    \item If $\mathbf{e}^i < \delta \min(\mathbf{e}^j), j\in \Omega$ and $\delta < 1.0$, \textit{continue}, else \textit{break}
    \item Update set of axis-aligned projections: $\Omega \leftarrow \Omega \cup i_{\alpha}$
    \item Solve the least squares optimization in (\ref{eqn:grassmann_ls}) and compute the residual on the Grassmannian: $\mathbf{R} = \mathbf{V} \mathbf{V}^T
      - \sum_{i \in \Omega} \beta_i\mathbf{Z}_i \mathbf{Z}_i^T$
  \item Reproject: Update $\mathbf{U}$ with two principal eigen vectors of  $\mathbf{R}$
  \end{enumerate}
  \end{enumerate}
\end{algorithm}

\noindent While the structural distortion metric is used to qualitatively evaluate the usefulness of the selected axis-aligned subspace in explaining the linear subspace, we estimate the residual subspace that could potentially contain information about the data, that is not described by the chosen axis-aligned subspace. This is carried out by first solving the following least squares optimization on the Grassmannian:
\begin{align}
\boldsymbol{{\beta}}_{\Omega} = \argmin_{\boldsymbol{\beta}} \|\mathbf{V} \mathbf{V}^T - \sum_{i \in \Omega} \beta_i \mathbf{Z}_i \mathbf{Z}_i^T\|_F^2 + \lambda \|\boldsymbol{\beta}\|_2,
\label{eqn:grassmann_ls}
\end{align} and estimating the residual subspace as shown in steps 2e and 2f (Algorithm \ref{alg:sparse_grass}) respectively. However, the residual subspace need not always promote the selection of another relevant axis-aligned subspace, particularly when the dimensions are inherently correlated. In such cases, redundant axis-aligned subspaces with similar structure could be chosen. To avoid that behavior, in step 2c, we compare the structural distortion for the chosen axis-aligned subspace to those picked so far and terminate if there is no improvement in the distortion. Note that for some linear projections $\mathbf{V}$, there exists no good axis-aligned projections and the residual $\mathbf{R}$ remains high or even increases as more $\mathbf{Z}_i$s are added. Note that this does not mean there is no axis-aligned subspace close to $\mathbf{V}$, but rather they do not preserve the structure of $\mathbf{V}$. To combat this challenge, we set an upper bound on the number of axis-aligned subspaces that can be chosen. This novel combination of axis-subspace selection and residual computation enables a robust and high-quality decomposition.


\begin{figure}[htbp]
\centering
\vspace{-2mm}
 \includegraphics[width=.9\linewidth]{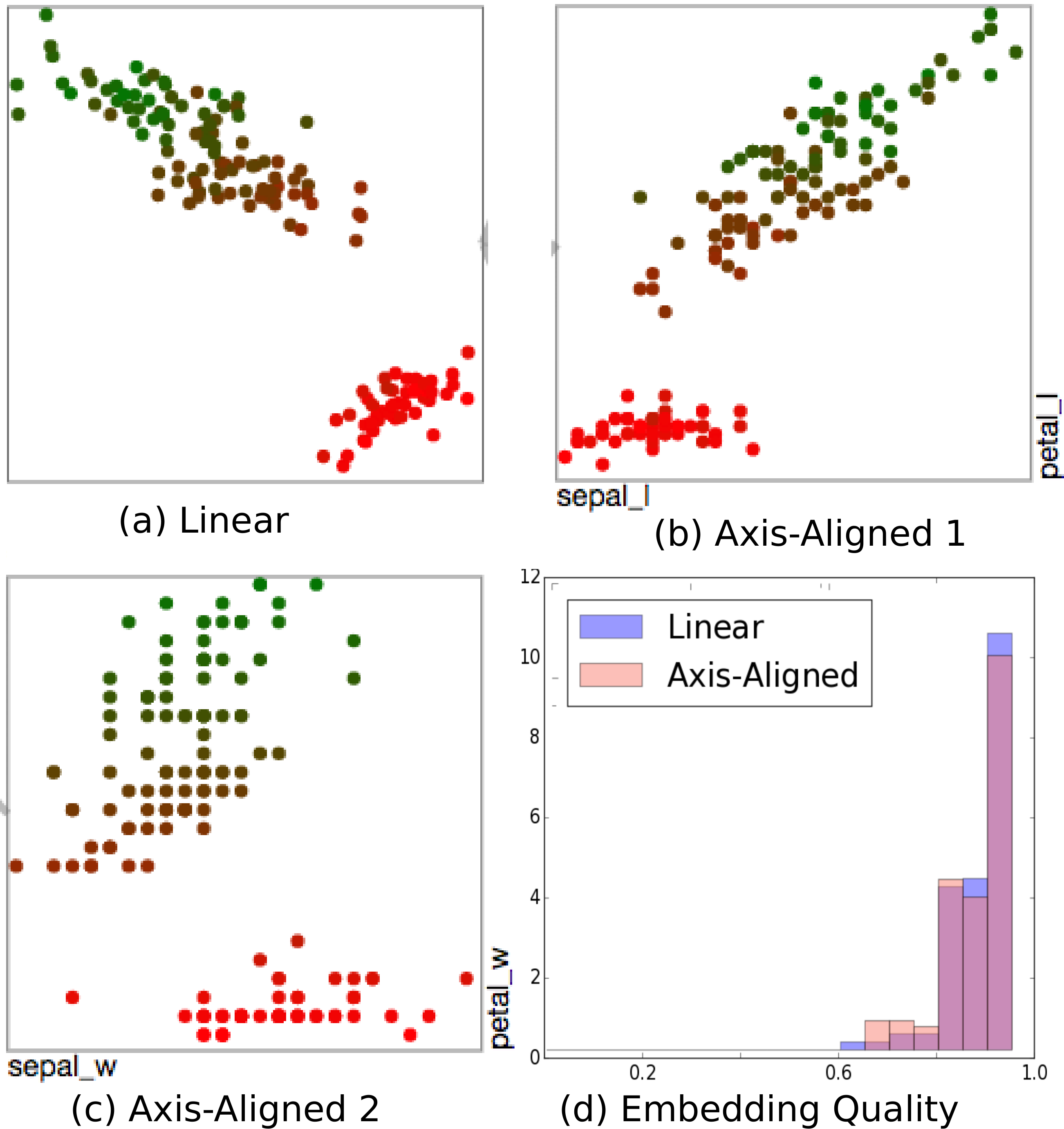}
 \caption{An example of the axis-aligned decomposition of a linear projection of iris dataset.}
\label{fig:decompositionExample}
\vspace{-5mm}
\end{figure}

\subsection{Example}
\label{sec:validation}
We use a simple example dataset to intuitively illustrate the decomposition
process. As shown in Figure~\ref{fig:decompositionExample}, we decompose
the LPP linear projection of the iris dataset (4D: sepal\_w, sepal\_l, petal\_w, petal\_l) 
into two structurally unique axis-aligned projections. 
As we can see, in Figure~\ref{fig:decompositionExample}(b)(c), each of
the axis-aligned projection uses two of unique dimensions in the 4D dataset. 

The compactness (i.e., avoid duplication) of the axis-aligned representation is one 
important goal of the proposed algorithm. 
There are likely many axis-aligned projections that contain similar patterns in the iris data, 
however, the proposed technique ensures that each axis-aligned projection captures
a unique structure. For the iris data example, in Figure~\ref{fig:decompositionExample}, 
the axis-aligned projection (b) is extracted first. Structurally, it is very close to the linear one. 
After the contribution of the axis-aligned projection (b) is removed from the linear projection (a) using Grassmann analysis, the second axis-aligned projection (c) reveals a different pattern. 
In addition to being compact, the decomposition preserves the
neighborhood structure from the linear projection with high fidelity. In order to demonstrate this, we compute the per-point precision-recall quality measure (see details in Section~\ref{sec:pr}, 
the higher the value the better) of the linear projection with respect to the high dimensional data. Further, we evaluate the aggregated per-point measures from the corresponding axis-aligned projections. For the aggregation, we use the maximum of the quality measure for each point across the different axis-aligned projection. As showed in Figure~\ref{fig:decompositionExample}(d), the histograms of the quality measure from both the cases highly overlap, indicating that for every point at least one of the axis-aligned
projections captures the neighborhood relationship observed in the linear projection.

\label{sec:method}
\section{Decomposition of Multiple Linear Projections}
\label{sec:multiLP}
In the previous section, we described our algorithm for identifying a concise set of axis-aligned projections that maximally describe the neighborhood structure observed in a given linear projection. However, as corroborated by several recent efforts~\cite{lehmann2016optimal,
LiuWangThiagarajan2015, wang2017subspace}, finding a diverse set of
representative projections is crucial for obtaining a comprehensive
understanding of high-dimensional data. Therefore, it is imperative to consider the scenario of obtaining a group of axis-aligned projections that jointly describe multiple linear projections of the data. 

However, decomposition of multiple linear projections presents additional challenges. First, finding a desirable set of representative linear projections is very hard. Even though many existing methods try to identify multiple interesting projections of the data, none of them explicitly optimize for both the diversity (i.e., cover various aspects of the data) and the
trustworthiness of the projection (i.e., make sure the projection
is not misleading). Ignoring either of the objectives may lead to undesirable
results.
Second, the decomposition of multiple linear projections entails the risk of redundancy, i.e., multiple axis-aligned projections may capture similar structure, and the challenge of aggregation, i.e., measuring the importance of an axis-aligned projection that is part of multiple linear projection decompositions.

\begin{figure}[htbp]
\centering
 \includegraphics[width=0.99\linewidth]{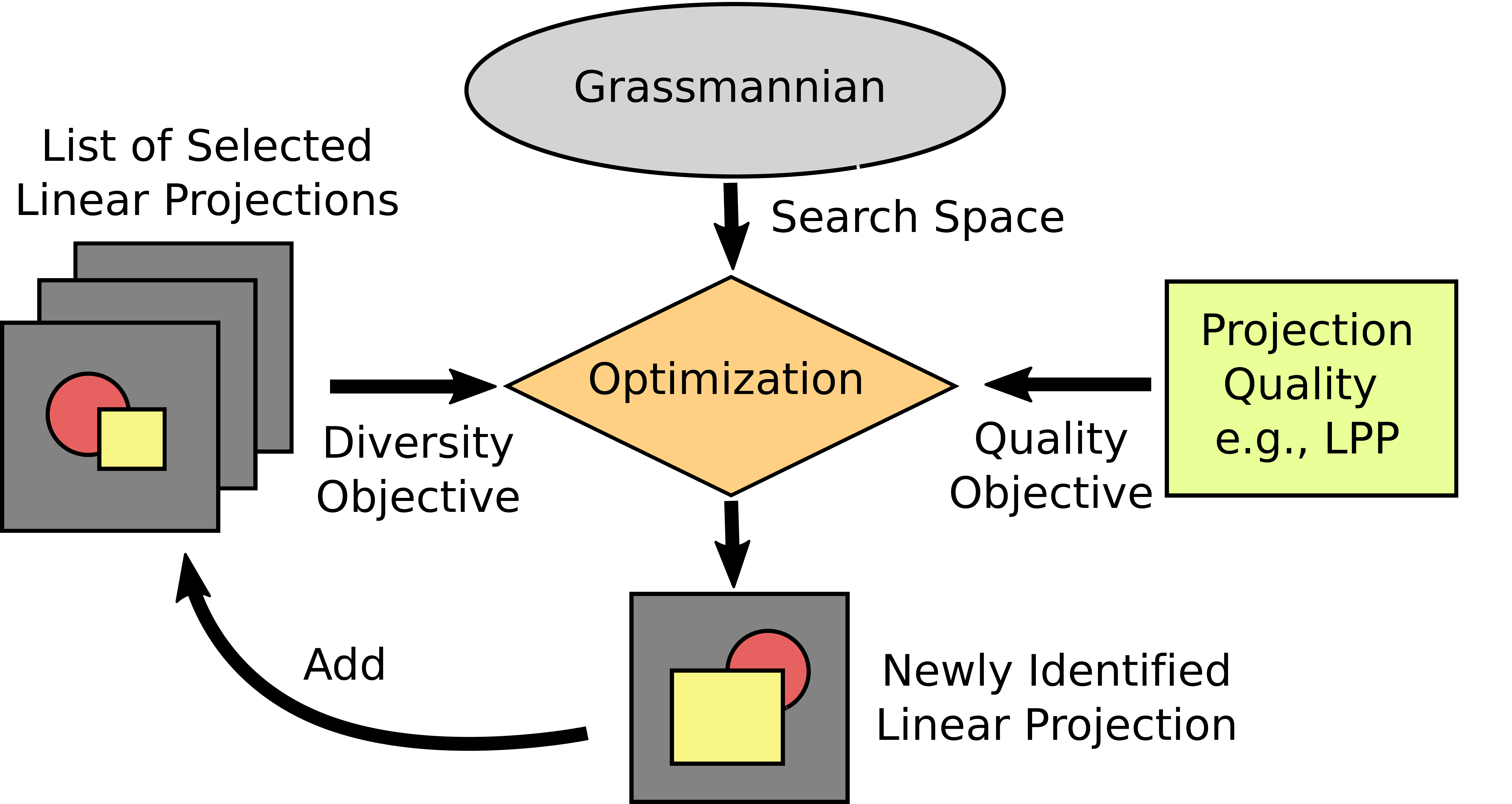}
 \caption{Illustration of the linear projection search algorithm.}
\label{fig:LPfinding}
\vspace{-5mm}
\end{figure}

In this section, we present our approach for finding multiple linear projections (see Figure~\ref{fig:LPfinding}) by jointly optimizing for embedding quality and diversity, using tools from Grassmann analysis. We generalize the algorithm in Section \ref{sec:method} by sequentially computing the decomposition for each of the linear projections. In this process, we avoid redundancy by promoting reuse of axis-aligned subspaces that were already chosen for another linear projection decomposition, thus producing a highly compact set of scatterplots for the user. Finally, we use tools from Dempster-Schafer theory to define an \textit{evidence} measure, which aggregates the contributions of each axis-aligned projection to the multiple linear projections.


%

\subsection{Finding Representative Linear Projections}
%

Given the data matrix $\mathbf{X} \in \mathbb{R}^{d \times n}$, our criterion for obtaining representative linear projections is that the inherent structure of the data points is preserved. This ranges from recovering the directions of maximal variance (PCA) to neighborhood structure (Locality Preserving Projections \cite{niyogi2004locality}) or class separation (Local Discriminant Embedding \cite{chen2005local}). Despite the varied nature of these techniques, all linear dimensionality reduction methods can be viewed through the lens of graph embedding. 

In this approach, we represent each vertex of a graph as a low-dimensional vector that preserves relationships between the vertex pairs, where the relationship is measured by a similarity metric that characterizes certain statistical or geometric properties of the data set. Let $G = \{\mathbf{X}, \mathbf{W}\}$ denote a undirected graph, where the matrix $\mathbf{W}\in\mathbb{R}^{n \times n}$ is the similarity matrix between all pairs of samples in $\mathbf{X}$. The Laplacian of the graph $G$ can be defined as $\mathbf{L} = \mathbf{D} - \mathbf{W}$, where $D_{ii} = \sum_{j \neq i} W_{ij}$. Denoting the linear projection by $\mathbf{V}$ and the corresponding embedding as $\mathbf{Y}=\mathbf{V}^T \mathbf{X}$, the problem of graph embedding can be written as: 
\begin{equation}
\min_{tr(\mathbf{V}^T\mathbf{X} \mathbf{B} \mathbf{X}^T \mathbf{V}) = \epsilon} tr(\mathbf{V}^T\mathbf{X} \mathbf{L} \mathbf{X}^T \mathbf{V}),
\label{eqn:lin_gemb}
\end{equation}where $tr(.)$ denotes the trace operator, the matrix $\mathbf{B}$ corresponds to the Laplacian of an optional penalty graph, typically used to regularize the learning, and $\epsilon$ is a penalty constraint. The solution to (\ref{eqn:lin_gemb}) can be obtained using generalized eigenvalue decomposition. Table \ref{table:formulation} lists the appropriate construction of the similarity graph and penalty graph Laplacians $\mathbf{L}$ and $\mathbf{B}$ for PCA, LPP and LDE.

\begin{table}[h]
	\centering
	\caption{Formulating linear dimensionality reduction techniques using the unified graph embedding framework in \cite{Yan2013}}
	\label{table:formulation}
	\begin{tabular}{|M{1cm}|M{3.5cm}|M{2.5cm}|}
		\hline
		\small
		\textbf{Method} & \textbf{Similarity Graph} & \textbf{Penalty Graph} \\ \hline
		PCA&       $W_{ij} = 1/n, i \neq j$  & $\mathbf{B} = \mathbf{I}$    \\ \hline
		LPP& $W_{ij} = \exp(-\gamma \|\mathbf{x}_i - \mathbf{x}_j\|^2)$, if $i \in \mathcal{N}(j)$ or $j \in  \mathcal{N}(i)$  & $\mathbf{B} = \mathbf{D}$ \\ \hline
		LDE&  $W_{ij} = 1, \text{ if } c_i = c_j$ & $\mathbf{B} = \mathbf{D}' - \mathbf{W}'$,  ${W}'_{ij}= 1, \text{ if } c_i \neq c_j$ \\ \hline
	\end{tabular}
\end{table}


However, since the intrinsic dimensionality of data is often greater than $2$, any 2D projection will invariably result in information loss. Hence, it is necessary to consider multiple 2D projections to obtain a more comprehensive view of the data. Similar to the approach in \cite{lehmann2016optimal}, we incorporate diversity as a quality measure to infer multiple projections. However, instead of comparing the structure of the two 2D projections, we propose to directly compare their corresponding subspaces on the Grassmannian manifold \cite{ye2014distance}, while ensuring the embedding quality is not entirely compromised. We use the squared chordal distance to compare two subspaces $\mathbf{V}_i$ and $\mathbf{V}_j$ on the Grassmannian:
\begin{align}
\nonumber
\rho^2(\mathbf{V}_{j},\mathbf{V}_{i}) &= 2-\|\mathbf{V}_i^T\mathbf{V}_{j}\|_F^2= 2-\text{Tr}\left(\mathbf{V}_{j}^T \mathbf{V}_i \mathbf{V}_i^T \mathbf{V}_{j}\right).
\end{align}

Our algorithm begins by inferring a linear projection for $\mathbf{X}$ by solving (\ref{eqn:lin_gemb}) for the desired embedding objective, e.g. LPP. Subsequently, we find the second projection that not only a provides good quality embedding but is also far from the first subspace. Assuming that we need to compute the $(j+1)^{\text{th}}$ subspace, its diversity is measured as the sum of distances between that subspace and all the previous $i = \{1, \ldots, j\}$ subspaces, i.e.,
\begin{align}
\nonumber
\sum_{i=1}^j \rho^2(\mathbf{V}_{j+1},\mathbf{V}_{i}) = j2-  \text{Tr}\left(\mathbf{V}_{j+1}^T \sum_{i=1}^j \left(\mathbf{V}_i \mathbf{V}_i^T\right) \mathbf{V}_{j+1}\right).
\end{align} Hence, the optimization problem for computing the $j+1^{\text{th}}$ subspace is
\begin{align}
\mathbf{V}_{j+1} = \argmin_{\mathbf{V}^T \mathbf{X} \mathbf{B} \mathbf{X}^T \mathbf{V} = \epsilon} \text{Tr}\left(\mathbf{V}^T \left( \mathbf{X} \mathbf{L} \mathbf{X}^T+ \alpha \sum_{i=1}^j \left( \mathbf{V}_i \mathbf{V}_i^T \right) \right) \mathbf{V}\right),
\label{eqn:graphemb_LE_incoh}
\end{align} where $\alpha$ is the trade-off parameter between embedding quality and the dissimilarity between subspaces. Since the comparison on the Grassmannian is independent of the data, in some cases, two different subspaces that are well separated can still produce 2D projections with similar structure. In order to avoid choosing redundant subspaces, we also verify if the actual projection can be obtained through a simple affine transformation of one of the previously found projections. 

\subsection{Decomposition}
Using the algorithm in Section \ref{sec:method}, we obtain the decomposition of each of the linear projections into a set of relevant axis-aligned subspaces. Since each of the linear projections is processed independently, there is a risk of redundancy, i.e. multiple axis-aligned projections with a similar structure can be picked. We avoid this by maintaining a global set $\Omega^G$ that contains all axis-aligned subspaces previously chosen for any of the linear projections. For the next linear projection, we choose a new axis-aligned projection over any of the projections in $\Omega^G$, only when its structural preservation property is superior.


\subsection{Inferring Evidence Measures}
\label{sec:evidence}
When considering the decomposition of multiple linear projections, the same axis-aligned subspace may contribute
to more than one linear subspace. Therefore, to evaluate
how much the axis-aligned subspace contributes to understanding the data on a whole, we propose to estimate \textit{evidence} scores based on structural distortions.

Dempster-Shafer theory (DST) is a general framework for reasoning with uncertainty \cite{shafer1976mathematical}, which we will utilize to understand the degree of belief of each axis-aligned subspaces in describing the data. Let $\Theta$ be the universal set of all hypotheses, i.e., the set of all axis-aligned subspaces in our case, and $2^{\Theta}$ be its power set. A probability mass can be assigned to every hypothesis $A \in 2^{\Theta}$ such that, $\mu(\varnothing) = 0, \sum_{A \in 2^{\Theta}} \mu(A) = 1,$ where $\varnothing$ denotes the empty set. This measure provides the confidence that hypothesis $A$ is true. Using DST, we can compute the uncertainty of the axis-aligned subspaces in representing a linear subspace using the belief function, $\sum_{B \subseteq A} \mu(B)$, which is the confidence on that hypothesis being supported by strong evidence. Using principles from DST, we can easily combine the evidence from multiple sources. In our case, this corresponds to combining beliefs of an axis-aligned subspace in describing multiple linear subspaces.

Given the set of axis-aligned subspaces $\Omega$ for a linear subspace $\mathbf{V}$, the mass corresponding to the axis-aligned subspace $\mathbf{Z}_i$, where $i \in \Omega$ is given as $\mu(\mathbf{Z}_i)$. This is estimated by computing the structural distortion $\mathbf{e}^i = \|\mathbf{C}\boldsymbol{\alpha} - \mathbf{b}^i\|_2$ and defining the mass $\eta$ to be the inverse of $\mathbf{e}^i$ defined as:
\begin{equation}
\eta = \eta_0 \left(1 - \frac{\mathbf{e}^i}{max(\{\mathbf{e}^j\}_{j\in \Omega})}\right)
\end{equation}Here $\eta_0$ is a parameter in the interval $[0,1]$ (chosen closer to $1$) that upper bounds the mass of any single hypothesis.


We apply Dempster's combination rule to combine beliefs of $\mathbf{Z}_i$ from all linear projections. Assuming that there are $P$ linear subspaces denoted by the orthonormal bases $\{\mathbf{V}_i\}_{i=1}^P$, the total mass can be accumulated as
\begin{equation}
\mu(\mathbf{Z}_i) = 1 - \prod_{p = 1}^P \left(1 - \eta_0 \left(1 - \frac{\mathbf{e}^i_p}{max(\{\mathbf{e}^j_p\}_{j\in \Omega, 0\leq p < P})}\right)\right).
\label{eqn:accbel}
\end{equation}Here $\mathbf{e}^i_p$ denotes the structural distortion obtained for the $p^{\text{th}}$ linear projection using the $i^{\text{th}}$ axis-aligned subspace. Finally, the normalized evidence measure of the axis-aligned subspace $\mathbf{Z}_i$ is given by
\begin{equation}
evid[i] = \frac{\mu(\mathbf{Z}_i)}{\max\left(\{\mu(\mathbf{Z}_j)\}_{j=1}^{|\mathcal{Q}|}\right)}, \forall i = 1, \cdots, |\mathcal{Q}|.
\label{eqn:evid}
\end{equation}

\begin{figure*}[!t]
\centering
\vspace{-2mm}
  \includegraphics[width=.96\linewidth]{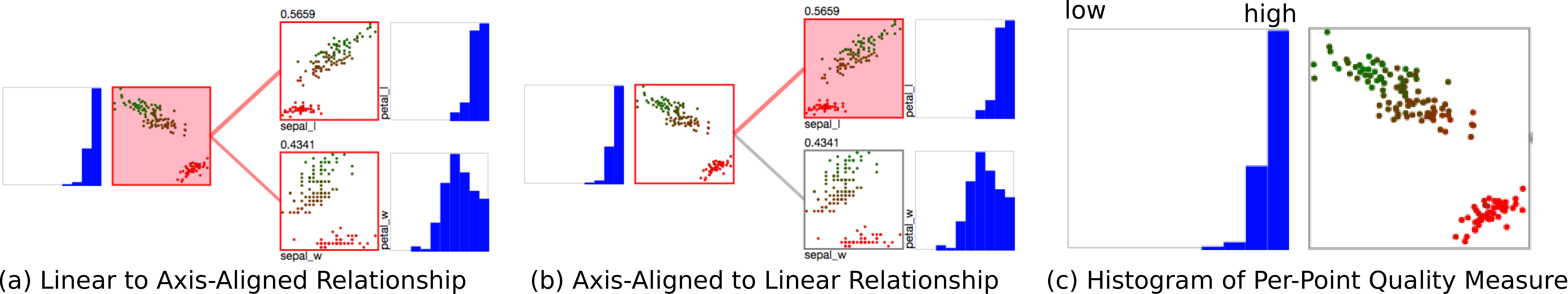}
 \caption{Projection Relationship View. As showed in (a)(b), the decomposition
 relationship is illustrated by a bipartite graph, the edges of the graph
 connect linear projections and their constituent axis-aligned projections from the proposed decomposition. 
 The currently selected projection is highlighted with a red background. As illustrated in (c), for each projection, the histogram of per-point precision-recall quality measure is shown for illustrating how well a given projection preserves the neighborhood structure of the full dimensional data.
 }
\label{fig:projRelationship}
\vspace{-5mm}
\end{figure*}

\section{Projections Relationship Visualization}
\label{sec:interface}


The previous sections discussed the computation pipeline that identifies
diverse, representative linear projections and subsequently decomposes them
into a compact set of axis-aligned projections. Though we can execute this pipeline directly as an offline analytic tool, visualizing the relationships between the linear and axis-aligned projections can not only present the computation results in a more accessible format but also help users develop
additional insights.
In this section, we discuss the design choice and the functionality for projection relationships visualization.

\noindent\textbf{Design Goal.}
To design an effective visual encoding, it is important to first identify
the exact information we want to communicate.
For projection relationships, the key information we want to convey
is the correspondence between a linear projection and its
axis-aligned decomposition. One interesting property is
the bi-directional nature of the relationships, i.e., a linear projection can be
decomposed into multiple axis-aligned projections, while the same axis-aligned
projection can also contribute to multiple linear projections.

Besides visualizing the connectivity among projections, we also want to help users develop a qualitative understanding of the structures captured in a linear projection via the decomposition into axis-aligned ones.
Therefore, it is critical to broadly understand the point-wise correlation, i.e., which part of the linear embedding corresponds to the axis-aligned one.
In particular, a meaningful animated transition from linear projection to
the axis-aligned projection can help highlight such a relationship.

To achieve these two design goals, we devised two visualization components:
the projection relationship view and the projection transition view.
The functionality and the rationale for the design choices are discussed in the following sections.

%

\noindent\textbf{Projection Relationship View.}
The projection relationship view provides an overview of the linear and
axis-aligned projections and encodes their decomposition relationships.
As discussed in Section~\ref{sec:multiLP}, the proposed computation method
first obtains the representative linear projections and then decomposes them into
axis-aligned ones. To express the two stages in the computation and visually
separate linear and axis-aligned projections, we encode the projections and their connections as a bipartite graph.

As illustrated in Figure~\ref{fig:teaser}(a), the projections
are encoded as nodes in the graph, where linear projections correspond to
nodes on the left column, and the axis-aligned projections are
nodes on the right column. Their decomposition relationship is expressed as edges in the bipartite graph.
A natural advantage of such an encoding is that the bi-directional nature of
the connectivity can be intuitively expressed.
As illustrated in Figure~\ref{fig:projRelationship}(a)(b), we show the
decomposition relationship from a linear (source) to axis-aligned projections
(targets),
as well as the aggregation relationship from an axis-aligned projection (source)
to the linear ones (targets) it describes. The source (also the actively
selected one) node are highlighted in light red and the edges from source
to targets are displayed as red lines.
In addition, the thickness of the line indicates the evidence contribution
(discussed in Section~\ref{sec:evidence}) of the axis-aligned projections for explaining the structure of the associated linear projection.
%
Finally, we allow the user to filter out axis-aligned projections 
(as well as the corresponding linear ones if all their decomposed axis-aligned projections are removed) if they have a very small evidence score (e.g., 0.05). 
This operation enables the user to focus on important relationships and projections,
and de-clutter the visualization space.

For each node, as illustrated in Figure~\ref{fig:projRelationship}(c),
we use the projection result as thumbnails to provide
a direct illustration of the configuration and structure of the point
embeddings.
However, the embedding alone may not inform the user how well the given
projection preserves the inherent neighborhood structure or class separation in high-dimensions. Hence, in order to obtain a qualitative understanding of the embeddings, for every projection, we attach a histogram plot of the per-point precision-recall quality measure (discussed later in this section), with respect to the high dimensional data. This will help users evaluate the amount of distortion in both linear and their constituent axis-aligned projections. For example, histograms concentrated to the far right are highly superior embeddings.

\begin{figure}[htbp]
\centering
\vspace{-2mm}
 \includegraphics[width=1.0\linewidth]{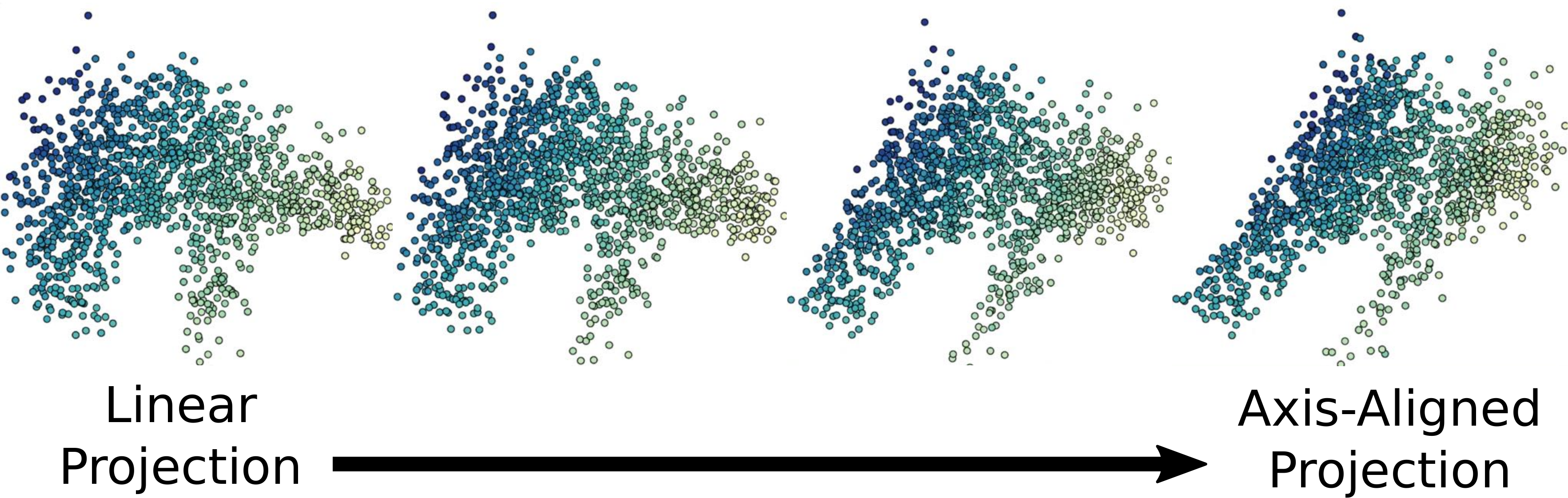}
 \caption{Animation projection transition from a linear projection to
 one of its constituent axis-aligned projections. As we can observe from
 the transition, the structure of the linear projection is well preserved
 in the axis-aligned one, which can be easily explained using the chosen variables.}
\label{fig:projTransition}
\vspace{-5mm}
\end{figure}

\noindent\textbf{Projection Transition View.}
The projection transition view (Figure~\ref{fig:teaser}(b)) illustrates
the point-wise correspondence between the linear and the axis-aligned projections through an animated transition between them.
The starting and ending points for the animation are always defined
by linear subspaces (axis-aligned is a special case of linear subspace). Consequently, in order to generate a meaningful transition, it is important to ensure that every frame in the animation corresponds to a valid, linear projection.
In this work, we utilize a linear projection matrix interpolation technique, similar to the one discussed in~\cite{BujaCookAsimov1997}.

Ultimately, we hope to utilize such a transition to connect the qualitative
insights we gain from the axis-aligned projections with the structure captured in the linear projection.
As illustrated in Figure~\ref{fig:projTransition}, as we can see the two
elongated clusters in the linear projection corresponds to a similar structure
in the axis-aligned projection that can be explained by two variables.

\noindent\textbf{Precision-Recall Measure.}
\label{sec:pr}
In order to provide an independent measure of accuracy to validate the quality
of any 2D projections, we employ a per-point precision-recall based measure (Figure~\ref{fig:projRelationship}(c)). 
The concept of precision and recall are widely used in information retrieval and
machine learning to evaluate false negative and false positive errors. 
While being independent of the optimization objective, this provides a natural trade-off between precision and recall metrics, 
which are both crucial for information visualization \cite{venna2010information}.
In our visualization, we measure precision and recall of the preserved
neighborhood for each data point in the projection with respect to the high dimensional data. We define the neighborhoods for a point $i$
in the original data and the visualization domain as, $\mathcal{N}_k (i)$, and
$\mathcal{N}'_{k'} (i)$ respectively where $k$ and $k'$ denote the number of points in the neighborhood. The precision and recall values are then computed as,
$\text{Precision} = \frac{|\mathcal{N}_k (i) \cap \mathcal{N}'_{k'} (i)|}{k'}, $
$\text{Recall} = \frac{|\mathcal{N}_k (i) \cap \mathcal{N}'_{k'} (i)|}{k}$.
For a given $k$ and $k'$ we estimate $\text{fidelity} = 0.5*\text{Precision} + 0.5*\text{Recall}$. For all results in this paper, we fixed both $k$ and $k'$ at $30$.

\section{Case Studies}
In the following sections, we apply the proposed technique to several
real word datasets to illustrate its effectiveness in capturing and
explaining structures in linear projections via their
axis-aligned decompositions.
The examples include both labeled and unlabeled data, with dimension ranging
between $7$ and $74$.

\begin{figure}[htbp]
\centering
  \includegraphics[width=.96\linewidth]{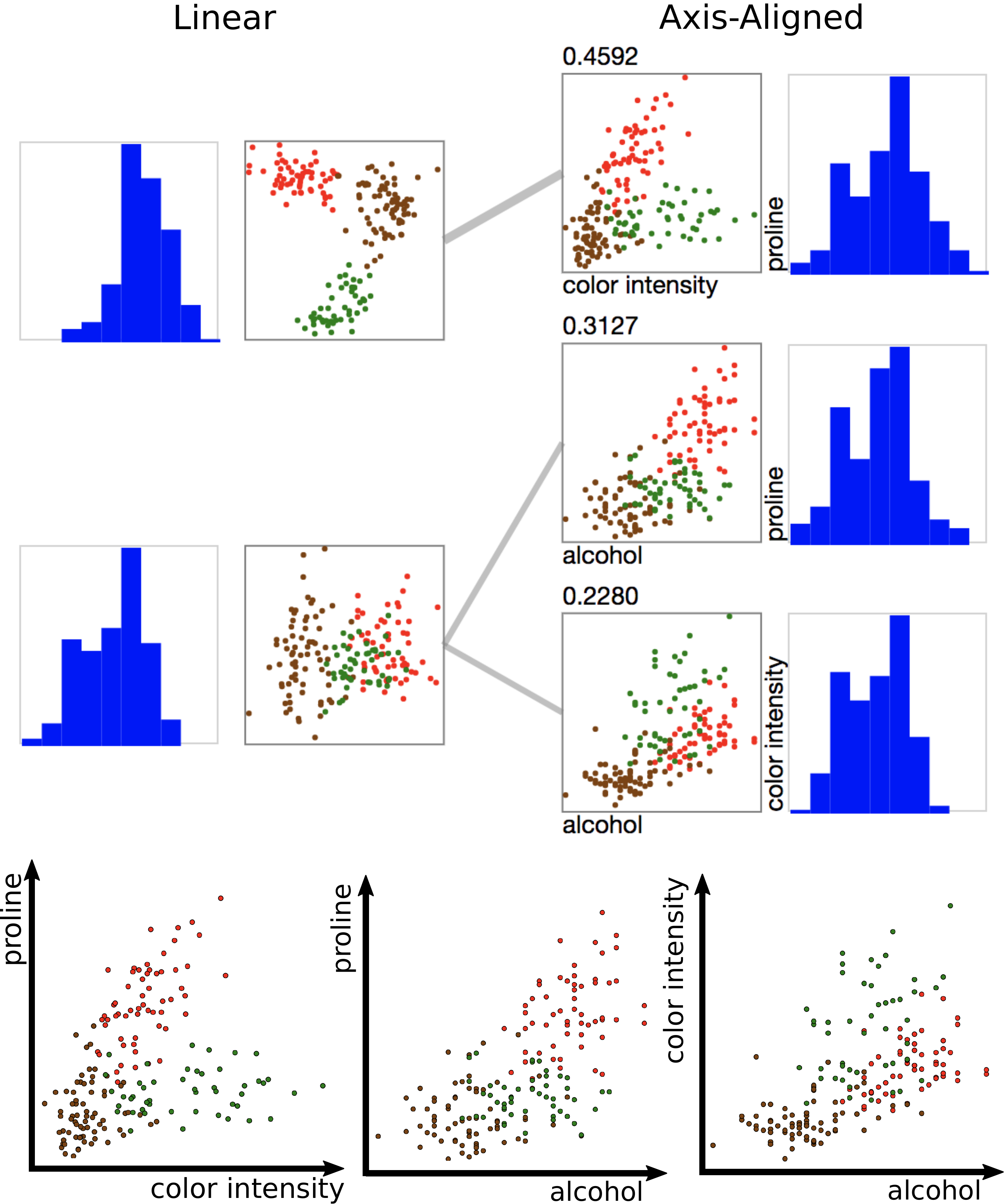}
 \caption{UCI Wine dataset. The red, brown and green classes correspond
 to three types of wines. The axis-aligned projections clearly illustrate
 which dimensions produce a class separation structure similar to the one
 observed in the diverse linear projections obtained with the LDE objective.}
\label{fig:wine}
\vspace{-0.1in}
\end{figure}

\subsection{Wine Dataset}
The UCI wine dataset is a commonly used dataset for testing
classification techniques. The data consists of 13 attributes
describing the characteristics of three types of wines (different
cultivars). The data has a well-defined class structure, therefore,
finding a good linear projection to reveal the class separation is not
a particularly challenging task. However, due to the complex linear
combination of the projection bases, interpreting the separation
structure in terms of individual data dimensions can be
difficult. Here, we illustrate how the proposed tool helps to address
this challenge via the axis-aligned decomposition of the linear
projections.

By utilizing the class separation objective from Local Discriminant
Embedding (LDE) (see Table~\ref{table:formulation}), as illustrated in
Figure~\ref{fig:wine}, we identify two linear projections that are
sufficiently different, yet still reasonably preserve the inherent
class structure.
%
As we can see in the first linear projection, the three classes are
well separated (highlighted by red, brown, green). However,
when examining the actual basis vectors of the projection,
they are dense with many active dimensions with similar coefficients.
As a result, even though the user can clearly see the
structure in the linear projection, it is almost impossible to
determine which attribute contributes to the class separation pattern.
By augmenting the linear projections with the relevant axis-aligned
ones, our tool provides a more intuitive view of the data. In
particular, as illustrated in Figure~\ref{fig:wine}, the axis-aligned
projections from the decomposition captures similar class separation
patterns. More specifically, we observe that the three axis-aligned
projections are formed by the attributes \emph{color intensity},
\emph{alcohol}, and \emph{proline}, which directly contribute to the
class separation structure. Interestingly, the three attributes also
define a 3D subspace where the clusters are cleanly separated.

\begin{figure}[htbp]
\centering
  \includegraphics[width=1.0\linewidth]{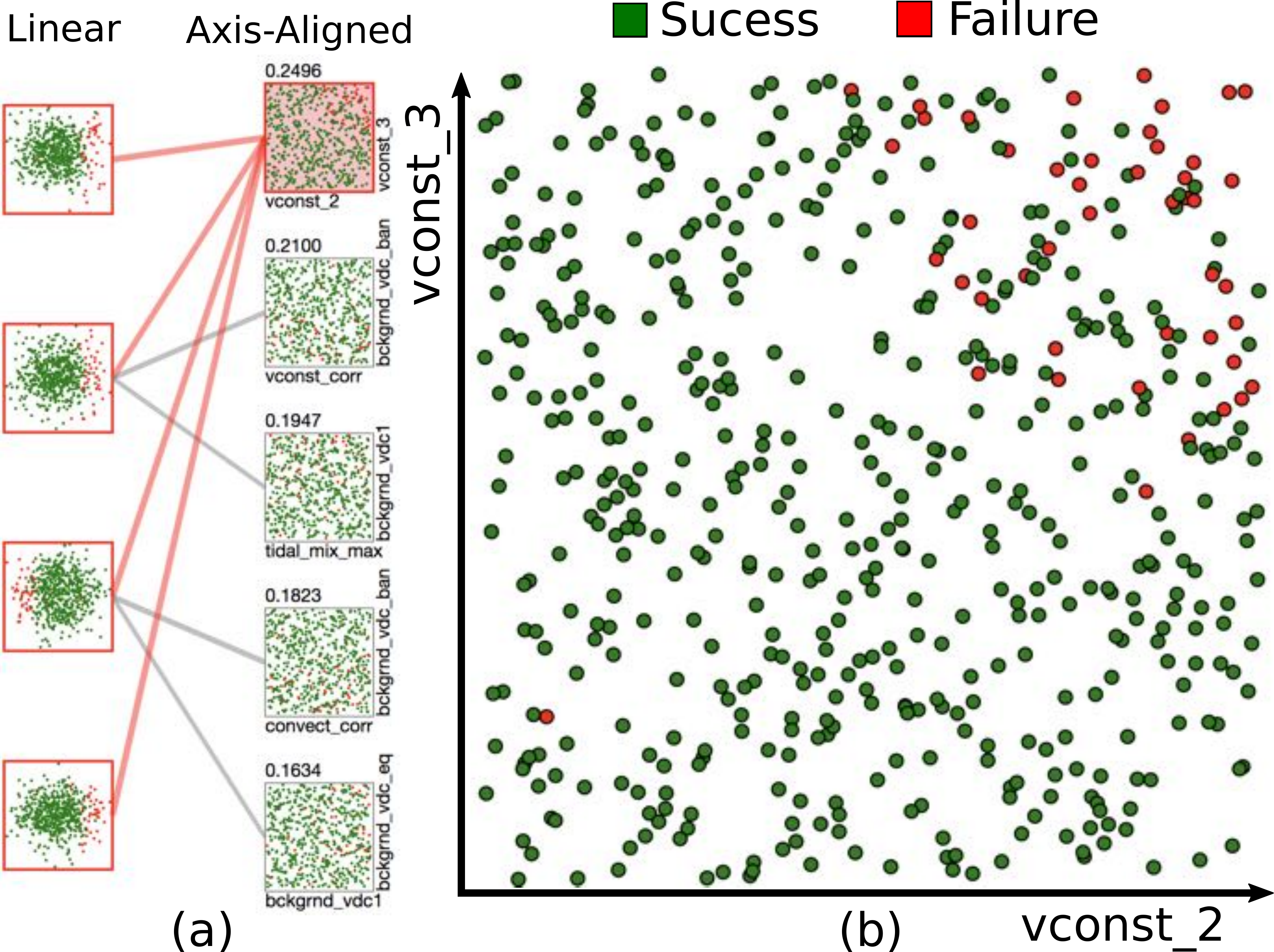}
 \caption{Climate Simulation Run Crashes. The red points correspond to the
 simulation crashes, and the green ones correspond to successful runs.}
\label{fig:climate}
\vspace{-0.1in}
\end{figure}

\begin{figure*}[!t]
\centering
  \includegraphics[width=.96\linewidth]{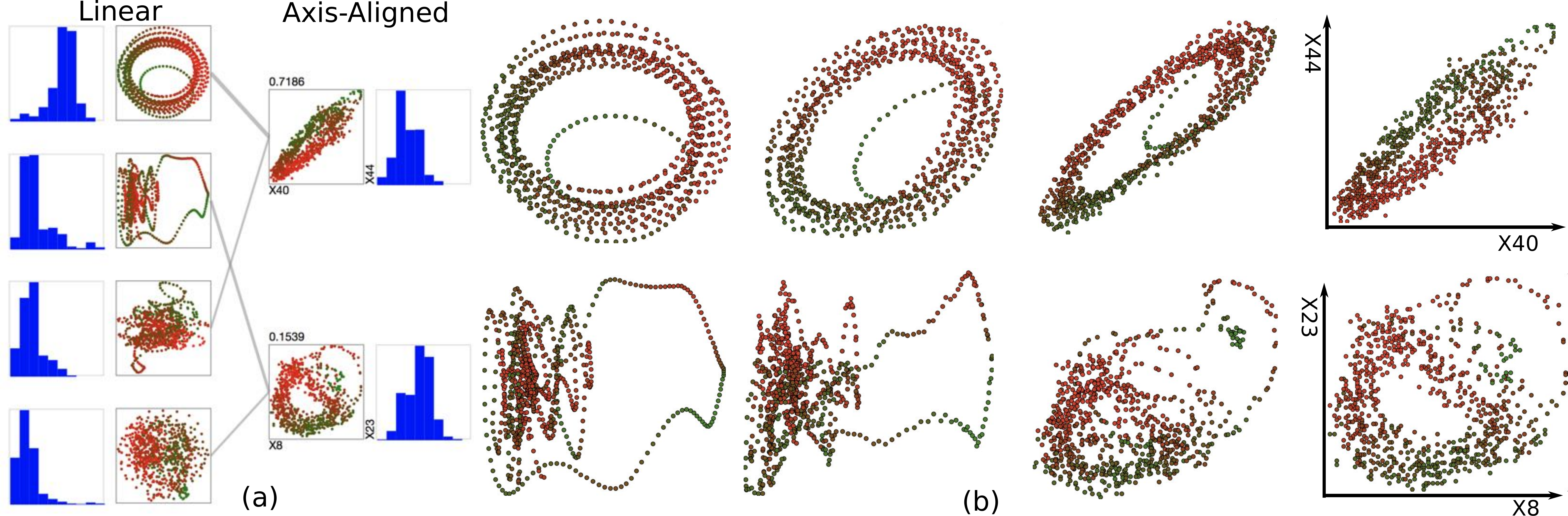}
 \caption{Seawater temperature forecasting dataset. The first axis-aligned projection
 captures the dominant periodic structure, while the second one
 highlights an additional loop that is different from one observed in the first projection.
 }
\label{fig:seawater}
\vspace{-0.1in}
\end{figure*}

\subsection{Climate Simulation Crashes Dataset}
An increasingly common approach to study the variability uncertainty
inherent in different climate models is to compute large ensembles of
simulations with varying input parameters. However, these studies
often include parameter combinations that are not well tested
or may even be inconsistent with each other. As a result, it is not
uncommon for the simulation to fail for a subset of runs. In this
situation, it is important to diagnose what parameter (combinations)
are involved in a crash to guide the debugging. Here we use a 
dataset which records successes and failures encountered during
simulations of the CCSM4 climate
model~\cite{LucasKleinTannahill2013}. The ensemble consists of 18
input parameters and $540$ latin hypercube samples $46$ of which
correspond to failures. The objective of the study is to find the
relationship between the parameter combinations and failure cases,
which can help determine the potential cause for the simulation
crashes.

As illustrated in Figure~\ref{fig:climate}(a), we utilize the LDE class
separation objective to produce the four representative linear projections,
in which the success and failure cases are clearly separated.
However, as before it is difficult to determine exactly which dimensions contribute
most to separating the success cases from failure cases.
We decompose these linear projections to verify if we can obtain a meaningful
separation in the axis-aligned projections, which can, in turn, reveal the direct impact of parameters on the simulation crashes.
As we can observe from the decomposition in
Figure~\ref{fig:climate}(a), despite the diversity of the linear
projections, we notice that all linear projections are effectively
described (evidenced by the edge thickness) by the highest ranked
axis-aligned projection. Note that, the ranking order is determined
based on the evidence scores obtained as discussed in
Section~\ref{sec:evidence}.
The same plot is enlarged in Figure~\ref{fig:climate}(b), which
reveals that the combination of high values for the attributes
\emph{vconst\_2}, \emph{vconst\_3} corresponds to all failure cases
(except one outlier) colored by red.  There is certainly some overlap
between red and green regions, which indicates that other factors
exist for exactly determining the outcome, thus justifying the need
for analyzing additional axis-aligned plots. However, we easily
identify the most useful axis-aligned projection from the exhaustive
set of $ \frac{18\times(18-1)}{2}$ combinations. Furthermore, it
indicates that the attributes \emph{vconst\_2} and \emph{vconst\_3}
attributes require the most attention when trying to figure out the
cause of crashes. Consequently, compared to linear projections, the
axis-aligned decomposition provides a much more direct and intuitive
interpretation of the patterns. Since the selection of the
axis-aligned projections is based on neighborhood structure preservation, the
other axis-aligned projections are selected due to their usefulness in
preserving the information about the success cases, even though they
do not reveal the class separation structure. The ability to produce
such complementary structure is the crucial aspect of our approach.

\subsection{Seawater Temperature Forecasting Dataset}
Time-series analysis and forecasting are required in many applications,
and in particular, long term prediction is very challenging. For this experiment,
we consider the sea water temperature forecasting dataset \cite{seawater},
which is a time series of weekly temperature measurements of sea water over
several years. Each data point is a time window of 52 weeks, which is shifted
one week forward for the next data point.  Altogether there are 823 data points
and 52 dimensions. The original goal of this data is to predict the future temperatures
based on previous time steps. For this analysis, we hope to identify the repeated
pattern, and more interestingly whether or not out-of-ordinary patterns exist in
the time series.

Note that, this moving-window representation of a time-series is
commonly referred as a delay embedding and is know to reveal periodic
structure in the form of loops. The periodic nature is thus very well
captured in the first linear projection obtained using LPP based
embedding optimization (Figure~\ref{fig:seawater}(a)). This projection
has high embedding quality when compared to rest of the projections,
as indicated by the quality measure histograms and it captures the
overall periodic pattern of the data.  The corresponding axis-aligned
projection, as illustrated in the top row of
Figure~\ref{fig:seawater}(b), shows a side view of the same pattern.
The second linear projection identifies a very interesting pattern. We
can notice that there is a second loop, different from the strong loop
found in the first linear projection. By viewing the transition from
this projection to its corresponding axis-aligned projection, new
insights about the structure is revealed. As shown in the bottom row
of Figure~\ref{fig:seawater}(b), the axis-aligned projection contains
an inner loop with most of the samples and an outer loop with a much
smaller number of points, thus revealing the presence of two different
periodic structures. In another word, our decomposition shows a
meaningful separation between the two dominant periodic structures,
which is critical information for understanding the complexity of this
prediction task.

\subsection{NIF Engineering Simulation Dataset}
The National Ignition Facility (NIF), a collaboration between Lawrence
Livermore, Los Alamos, and Sandia National Laboratories as well as The
University of Rochester and General Atomics, is aimed at demonstrating
inertial confinement fusion (ICF), that is, thermonuclear ignition and
energy gain in a laboratory setting. Fundamentally, the goal of NIF is
to search the parameter space to find the region that leads to
near-optimal performance, in terms of the energy yield. The dataset
considered here is a so-called engineering or macro-physics simulation
ensemble in which an implosion is simulated using various different in
parameters, such as, laser power, pulse shape etc.. From these
simulations scientists extract a set of \emph{drivers}, physical
quantities thought to determine the behavior of the resulting
implosion. These drivers are then analyzed with respect to the energy
yield to better understand how to optimize future experiments. The
dataset, we consider for our the analysis consists of $1304$ samples
with $6$ drivers: down scatter fraction (dsf), peak velocity (pv),
entropy (sument), (totrhorba), pressure at the centre (prcent), and
hotspot radius (hsrad).

\begin{figure}[htbp]
\centering
  \includegraphics[width=.96\linewidth]{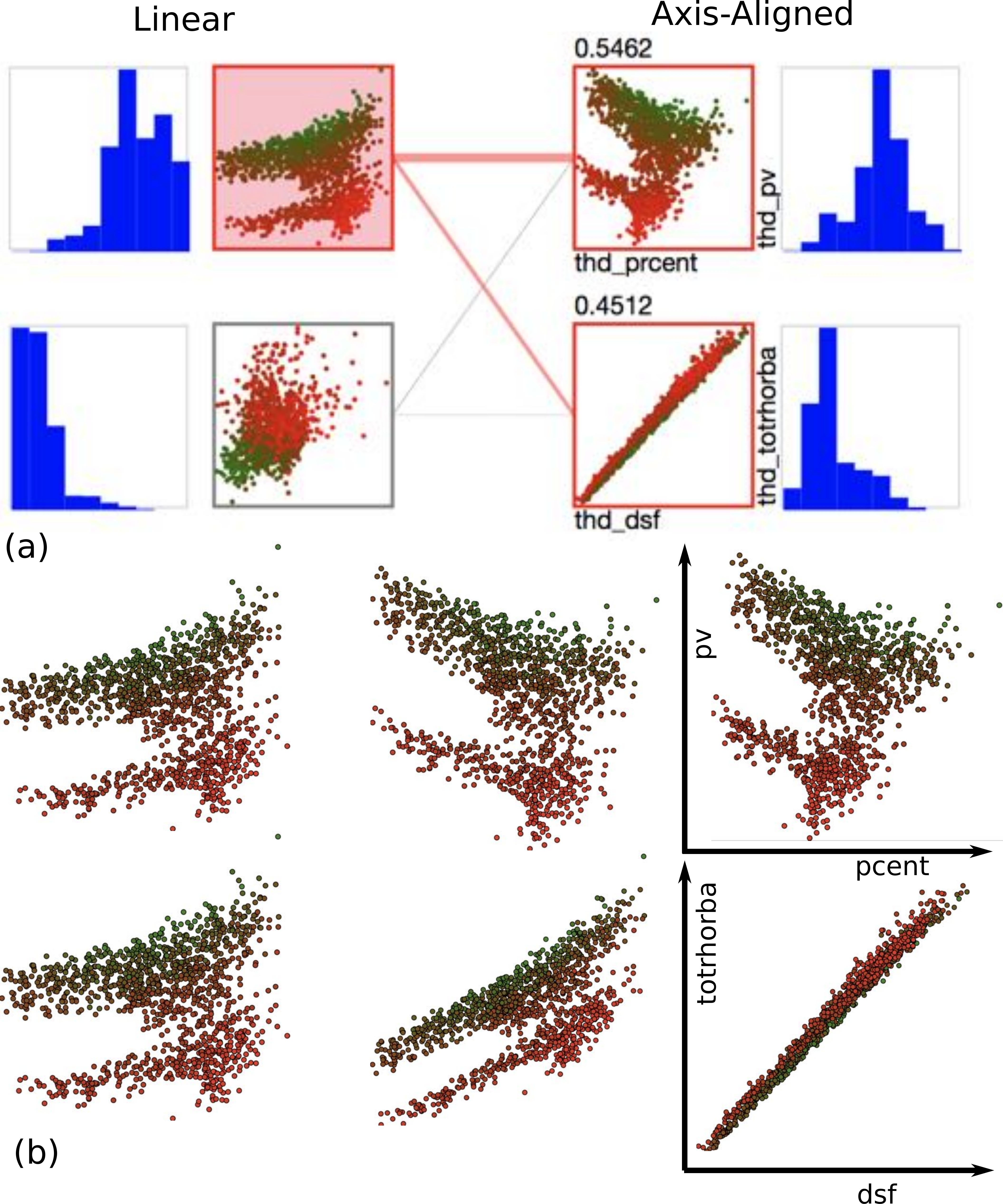}
 \caption{NIF Simulation Data. Our proposed approach reveals the
 presence of two true clusters in the attribute \textit{pv} and a strong
 correlation between the attributes \textit{dsf} and \textit{totrhorba},
 though the projection artifacts mask the neighborhood structure.}
\label{fig:nif}
\vspace{-0.1in}
\end{figure}

For this exploratory analysis, we choose the LPP objective to find the
diverse set of representative linear
projections. Figure~\ref{fig:nif}(a) shows the two representative
linear projections produced by our approach, and the two axis-aligned
projections that describe the structures induced in the linear
embeddings. The projection with the highest evidence is comprised of
the attributes \emph{pv}, \emph{pcent}, which reveals the two
elongated protruding clusters.  The transition from the linear
projection to the \{\emph{pv}, \emph{pcent}\} subspace (top row of
Figure~\ref{fig:nif}(b)) illustrates the small structural differences
between them. As we can see in the axis-aligned plot, the high/low
value of \emph{pv} roughly corresponds to two protruding clusters.
The second highest projection reveals a strong linear correlation
between \emph{totrhorba} and \emph{dsf}.  Interestingly, the
precision-recall based quality measure indicates that the second
axis-aligned projection contains significantly more artifacts than the
first one, yet the evidence measure claims both axis-aligned projections
capture important structures.  To help explain the cause of the
projection error, we apply a transition between the two projections
(bottom row of Figure~\ref{fig:nif}(b)).  Here we notice that the two
clusters in the original linear projection become overlapped while
transitioning to the \{\emph{totrhorba},\emph{hsrad}\} plot.
Consequently, we infer that there are two
geometrically distinct regions in the high-dimensional space where the
attributes \emph{totrhorba}, \emph{dsf} are strongly correlated.
Due to the overlap in the axis-aligned projection,
the neighborhood structure is partially lost (as indicated by the histogram of the quality).
Nevertheless, our algorithm is able to correctly determine that this
relationship is valid and assigns a high evidence score.

Consulting with the relevant physicists, we have confirmed that the
correlation observed between the two attributes agrees with the
underlying physics. More importantly, the axis-aligned projection that
captures the two protruding clusters in the linear projection are very
useful, since that structure is inherently more challenging to
interpret in a linear projection as stated by the physicists in a previous study (part of the
motivation for this work).

\section{Discussion}
This work introduced a novel algorithm for decomposing structure-preserving linear projections into a compact set of interpretable axis-aligned scatterplots. Combined with a novel optimization technique for generating representative linear projections and an intuitive visual interface, we allow users to explore complex high-dimensional data and make connections between the observed structure (e.g., clusters, correlations) and the actual data dimensions.
By jointly examining the structure preservation effectiveness of linear projections via the quality measure histograms 
and the evidence of the axis-aligned plots via the edge thicknesses in the relationship view, the users can obtain a good understanding of how much of the inherent high-dimensional structure can be explained by the data dimensions. 

As demonstrated by our case studies, the proposed method is widely applicable to
a broad range of unsupervised and supervised analysis tasks. A potential limitation of this approach is that the axis-aligned plots are not easily comparable to their corresponding linear projections, when the attributes are categorical (binary or multiple states) in nature. In our case studies, we are dropping the categorical variables out of the analysis, if they exist. 
Due to the efficiency of the proposed algorithm, our method can adapt to both
very high dimensions (>100), as well as large sample size (>10000).
The dominant complexity of the linear projection finding step arises from the generalized eigenvalue decomposition of the matrices of size $d \times d$, where $d$ is the number of dimensions. On the other hand, the greedy algorithm for performing the decomposition incurs a complexity of order $\mathcal{O}(dn)$, where $n$ is the total number of samples. 
%



%

\bibliographystyle{eg-alpha-doi}

\bibliography{egbibsample}
\end{document}